\pdfoutput=1
\documentclass[11pt]{article}

\usepackage{naacl2021}

\usepackage{times}
\usepackage{latexsym}

\usepackage[T1]{fontenc}
\usepackage[utf8]{inputenc}

\usepackage{microtype}

\usepackage{multirow}
\usepackage{CJKutf8}
\usepackage{amsmath}
\usepackage{xspace}
\usepackage{graphicx}
\usepackage[switch]{lineno}
\usepackage{enumitem}
\usepackage{paralist}
\usepackage{caption}
\usepackage{subcaption}
\title{Autocorrect in the Process of Translation --- Multi-task Learning Improves Dialogue Machine Translation}

\author{
    Tao Wang\textsuperscript{1,2}, Chengqi Zhao\textsuperscript{1}, Mingxuan Wang\textsuperscript{1}, Lei Li\textsuperscript{1},  Deyi Xiong\textsuperscript{3}\thanks{\ \ Corresponding author.} \\
    \textsuperscript{1}ByteDance AI Lab \\
    \textsuperscript{2}School of Computer Science and Technology, Soochow University, Suzhou, China \\
    \textsuperscript{3}College of Intelligence and Computing, Tianjin University, Tianjin, China \\
    \texttt{\{wangtao.960826, zhaochengqi.d, wangmingxuan.89\}@bytedance.com} \\
    \texttt{\{lilei.02\}@bytedance.com} \\
    \texttt{dyxiong@tju.edu.cn}
}

\newcommand{\method}{\textsc{NMTdial}\xspace}
\newcommand{\mbase}{\textsc{BASE}\xspace}
\newcommand{\mrepair}{\textsc{REPAIRdial}\xspace}
\newcommand{\mrobust}{\textsc{ROBUSTdial}\xspace}
\newcommand{\mmlt}{\textsc{MTLdial}\xspace}

\newcommand{\droppro}{\texttt{ProDrop}\xspace}
\newcommand{\droppun}{\texttt{PunDrop}\xspace}
\newcommand{\typo}{\texttt{DialTypo}\xspace}

\begin{document}
% \linenumbers
\begin{CJK}{UTF8}{gbsn}
\maketitle

\begin{abstract}
Automatic translation of dialogue texts is a much needed demand in many real life scenarios. However, current neural machine translation systems usually deliver  unsatisfying translation results of dialogue texts.
In this paper, we conduct a deep analysis of a dialogue corpus and summarize three major issues on dialogue translation, including pronoun dropping (\droppro), punctuation dropping (\droppun), and typos (\typo).
In response to these challenges, we propose a joint learning method to identify omission and typo in the process of translating, and utilize context to translate dialogue utterances.
%The proposed model will automatically learn to fill in and correct while learning how to translate.
To properly evaluate the performance, we propose a manually annotated dataset with 1,931 Chinese-English parallel utterances from 300 dialogues as a benchmark testbed for dialogue translation.
Our experiments show that the proposed method improves translation quality by 3.2 BLEU over the baselines. It also elevates the recovery rate of omitted pronouns from 26.09\% to 47.16\%.
The code and dataset are publicly available at https://github.com/rgwt123/DialogueMT.
\end{abstract}

\section{Introduction}
\label{sec:intro}
Remarkable progress has been made in Neural Machine Translation (NMT) \cite{bahdanau2015neural,wu2016google,lin2020pre,liu2020multilingual} in recent years, which has been widely applied in everyday life.
A typical scenario for such application is translating dialogue texts, in particular the record of group chats or movie subtitles, 
% which assists people from different languages with cross-language chat and comprehension capabilities. 
which helps people of different languages understand cross-language chat and improve their comprehension capabilities.

\begin{table}[]
\centering
\footnotesize
\begin{tabular}{l|l}
\hline
(1) & \begin{tabular}[c]{@{}l@{}}Nancy怎么了?\\ ${[}$她${]}_{drop}$是不是哭了啊。\end{tabular}              \\ \hline
MT  & \begin{tabular}[c]{@{}l@{}}What happened to Nancy? \\ Did you cry?\end{tabular} \\ \hline
REF & \begin{tabular}[c]{@{}l@{}}What happened to Nancy? \\ Did she cry?\end{tabular} \\ \hline\hline
(2) & Nancy怎么了${[}$?${]}_{drop}$是不是哭了啊。               \\ \hline
MT  & Did Nancy cry?                       \\ \hline
REF & What happened to Nancy? Did she cry? \\ \hline\hline
(3) & Nancy怎么${[}$乐${]}_{typo}$?                      \\ \hline
MT  & How happy is Nancy?                  \\ \hline
REF & What happened to Nancy?              \\ \hline
\end{tabular}
\caption{Examples of \droppro (1), \droppun (2) and \typo (3). MT is translation results from Google Translate while REF is references.}
\label{table_3errors}
\end{table}

However, traditional NMT models translate texts in a sentence-by-sentence manner and focus on the formal text input, such as WMT news translation \cite{barrault2020findings}, while the translation of dialogue must take the meaning of context and the input noise into account.
Table \ref{table_3errors} shows examples of dialogue fragment in Chinese and their translation in English. 
Example (1) demonstrates that the omission in traditional translation  (e.g., dropped pronouns in Chinese) leads to inaccurate translation results. 
% For instance, the dropped pronoun ``她/she'' leads to bad-quality machine translation results, which do not properly convey the original meaning of Chinese sentences.
% Similarly, the omission of punctuation (a common issue in online chat) behind the word ``是" in the fifth segment changes the way that the segment is comprehended and therefore the way that it is translated. 

% Despite of its huge potential application, dialogue translation has so far not been widely explored.
Despite its vast potential application, efforts of exploration into dialogue translation are far from enough.
Existing works \cite{wang2016automatic, maruf2018contextual} focus on either extracting dialogues from parallel corpora, such as OpenSubtitles \cite{lison2019open}, or leveraging speaker information for integrating dialogue context into neural models. 
% However, specific phenomena and challenges in dialogue translation are not deeply investigated. %Additionally, dialogue translation models still suffer from the lack of both training data and benchmark testsets.
Also, the lack of both training data and benchmark test set makes current dialogue translation models far from satisfying and need to be further improved.

In this paper, we try to alleviate the afore-mentioned challenges in dialogue translation. We first analyze a fraction of a dialogue corpus and summarize three critical issues in dialogue translation, including \droppro, \droppun, and \typo.
% 呼应题目 后面的DialMTL和DialNMT在
% Then we propose a Multi-Task Learning (\mmlt) approach that jointly learns to repair and translate at the same time.
Then we design a Multi-Task Learning (\mmlt) approach that learns to self-correct sentences in the process of translating.
The model's encoder part automatically learns how to de-noise the noise input via explicit supervisory signals provided by additional contextual labeling.
% 模型的encoder部分通过额外的labeling的显式监督信号，从输入中自动学习到filling和correcting的信息，有助于更好的翻译。
We also propose three strong baselines for dialogue translation, including repair (\mrepair) and robust (\mrobust) model.
% 为了缓解对话数据稀缺带来的挑战，我们使用双语平行语料中的sub-documents来让模型学习和使用cross-sentence context信息，而不需要任何对话双语数据。
To alleviate the challenges arising from the scarcity of dialogue data, we use sub-documents in the bilingual parallel corpus to enable the model to learn from cross-sentence context.

Additionally as for evaluation, the most commonly used BLEU metric \cite{Papineni2001BleuAM} for NMT is not good enough to provide a deep look into the translation quality in such a scenario.
% Thus, we build a Chinese-English test set containing sentences with the \droppro, \droppun and \typo phenomena with their human translations and annotations.
Thus, we build a Chinese-English test set containing sentences with the issues in \droppro, \droppun and \typo, attached with the human translation and annotation.
%The purpose of building this test set is to establish a benchmark  for evaluating dialogue  translation with linguistic and dialog-translation-oriented metrics in addition to traditional automatic metrics. 
Finally, we get a test set of 300 dialogues with 1,931 parallel sentences.
% We will provide the details of the way that we design and build this test set in the fourth Section. 
%We will provide the details of how we design and build this test set in the fourth Section.

The main contributions of this paper are as follows:
\begin{inparaenum}[a)]
    \item We analyze  three challenges  \droppro, \droppun and \typo, which greatly impact the understanding and translation of a dialogue.
    \item We propose a contextual multi-task learning method to tackle the analyzed challenges.
    \item We create a Chinese-English test set specifically containing those problems and conduct experiments to evaluate proposed method on this test set.
\end{inparaenum}

\section{Analysis on Dialogue Translation}
\label{sec:analysis}
% \begin{figure}[t]
% \centering
% \includegraphics[scale=0.8]{pdf_errors.pdf}
% \caption
% {Examples of \droppro (1), \droppun (2), \typo (3), wrong segmentation (4) and common translation errors (5). English translations in red/blue is wrong/correct. }
% \label{pdf_4errors}
% \end{figure}

% \begin{figure*}[t]
% %\setlength{\belowcaptionskip}{-0.6cm} 
% \centering
% \includegraphics[scale=0.9]{pdf_model.pdf}
% \caption
% {Overall diagram of \method. The left part demonstrates the process of data generation, and the right part displays the three proposed methods. ①/②/③ represent \mrepair, \mrobust and \mmlt respectively.}
% \label{pdf_model}
% \end{figure*}

\begin{figure*}[htbp]
\centering
\begin{subfigure}[b]{0.45\textwidth}               
\centering                                                \includegraphics[width=\textwidth]{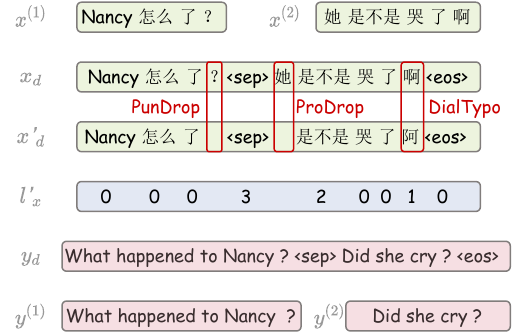}
\caption{}
\label{pdf_model_a}
\end{subfigure}
\begin{subfigure}[b]{0.45\textwidth}   
%第二张子图
\centering                      
\includegraphics[width=\textwidth]{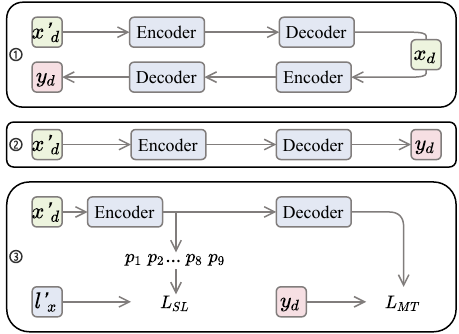}
\caption{}
\label{pdf_model_b}
\end{subfigure}
\caption
{Overall diagram of \method. (a) demonstrates the process of data generation, and (b) displays the three proposed methods. ①/②/③ represent \mrepair, \mrobust and \mmlt respectively.}
\label{pdf_model}
\end{figure*}

% 为了研究对话翻译中容易出现的问题，我们需要一个有着对话文本以及对应翻译的语料，
%We begin with a study on a bilingual dialogue corpus.\footnote{https://github.com/longyuewangdcu/tvsub}
%This corpus contains more than two million Chinese-English sentence pairs extracted from the television episodes. It is designed for dialogue translation with wide context information.

% 之前有不少对翻译错误进行人工分析的工作，特别是篇章翻译领域。Elena Voita研究了english-russian翻译，并指出了deixis，ellipsis，lexical cohesion三个主要问题。
There were already some manual analyses of translation errors, especially in the field of discourse translation.
\citet{voita2019good} study English-Russian translation and find three main challenges for discourse translation: deixis, ellipsis, and lexical cohesion.
% 对中英翻译来说，tense consistency，conjunction presence是两个经常被提起的问题。
For Chinese-English translation, tense consistency, connective mismatch, and content-heavy sentences are the most common issues \cite{li2014assessing}.

Different from previous works, we mainly analyze the specific phenomena in dialogue translation. We begin with a study on a bilingual dialogue corpus \cite{wang2018translating}.\footnote{https://github.com/longyuewangdcu/tvsub} We translate source sentences into the target language at sentence level and compare translation results with reference at dialogue level.
Around 1,000 dialogues are evaluated, and the results are reported in Table \ref{table_statistic-of-errors}. 
From the statistic, we observe two persistent dialogue translation problems: pronoun dropping (\droppro),  punctuation dropping(\droppun).
% 这和我们平时在真实IM聊天场景下收集的错误基本类似，除了错别字的错误。这是因为对话语料经过了校对，去除了错别字，所以我们在下面也会介绍这个问题。
The phenomenon is consistent with the issue we collect in practical Instant Messaging (IM) chat scenarios, except for typos since the analyzed dialogue corpus has been proofread to remove typos. 
% However, typos are very common in real-world scenarios. Therefore, we include this type of errors in our analysis too.
% Next we will discuss the most serious problems in dialogue which affect the quality of machine translation system. 
% Next we will discuss the most serious problems in dialogue translation which affect the quality of translation results generated by the machine translation system.
% There are also some other common translation problems which is not our key point.
% There are also some other common translation problems, but as they are not our key focus, we will not present them in detail in this paper.

\begin{table}[]
\centering
\small
\begin{tabular}{|l|c|}
\hline
\textbf{Types of phenomena}    & \textbf{Frequency} \\ \hline
\multicolumn{1}{|l|}{Correct} & 88.1\%             \\
%\multicolumn{1}{|l|}{Incorrect:}  &                    \\
\droppro               & 4.3\%              \\
\droppun           & 3.2\%              \\
Incorrect segmentation     & 2.4\%              \\
Other translation errors        & 2.0\%                \\ \hline
\end{tabular}
\caption{Manual evaluation of dialogue samples.}
\label{table_statistic-of-errors}
\end{table}

\subsection{Pronoun Dropping}
Pronouns are frequently omitted in pro-drop languages \cite{huang1989pro}, such as Chinese, Japanese, Korean, Vietnamese, and Slavic languages.
% This phenomenon is more serious in dialogue, where pronouns are usually omitted as long as the interlocutors have consensus on them.
Such phenomenon are more frequent in dialogue, where the interlocutors are both aware of what's omitted in the context.
However, when translating a pro-drop language into a non-pro-drop language (e.g., English)\footnote{https://en.wikipedia.org/wiki/Pro-drop\_language}, it is hard to translate those omitted pronouns, % resulting in grammatical errors or semantic deficiencies in the target language.
resulting in grammatical errors or semantic inaccuracies in the target language.
The first conversation in Table \ref{table_3errors} is an example.
%The subject ``她/she'' is  omitted for brevity in Chinese dialogue,
% while it is not very common to drop the subject of English sentences.
%while it is not very common to drop the subject in an English sentence.
%In this case, the machine translation system tends to auto-complete the results as the most common expression ``What happened to Nancy? Did you cry''~\footnote{The translation result is from Google Translate: https://translate.google.com/}.

\subsection{Punctuation Dropping}
In dialogue scenarios, such as IM software, punctuation is often omitted and  users tend to segment sentences with spaces. The problem becomes much serious in languages with no spaces, such as Chinese, Japanese, Korean, and Thai. 
Table \ref{table_3errors} shows this phenomenon in Example (2).
% A heuristic strategy is to directly replace the spaces with punctuation. 
% However, it will introduce additional noise in practical scenarios. For example, Chinese is mixed with English, numbers, etc., or spaces are used to highlight keywords instead of separating sentences. 
% However, it will introduce additional problems in practical scenarios like Chinese mixed with English, numbers, etc., or spaces used to highlight keywords instead of separating sentences.

\subsection{Dialogue Typos}
% 错别字检测被广泛的应用在各种任务中，比如说OCR，文本搜索等。它是一个重要却又十分有挑战性的问题。在对话翻译的场景下，错别字现象往往比正式文本出现的概率会更高。图1中的（3）是一个典型的场景。幸亏的幸字被打成了发音以及笔形都十分相似的辛字，这会导致分词后辛亏不在词表中，或者被BPE更加细粒度得划分为，完全丢失原来的语义。
Typo repairing is another fundamental but very challenging practical problem. 
In dialogue translation, typos or misspellings are very common, 
% which dramatically affects the quality of machine translation system. 
which dramatically undermine the quality of translation output produced by machine translation.
Table \ref{table_3errors} shows this phenomenon in Example (3).

\section{Approach to \method}
\label{sec:approach}
This section aims to propose a unified framework that facilitates NMT to  correct noisy inputs in dialogue neural machine translation (\method). 
% And we propose three different methods, which are \mrepair, \mrobust and \mmlt.
The framework includes three different methods, which are \mrepair, \mrobust and \mmlt. 

\subsection{Contextual Perturbation Example Generation}
The most challenging problem for \method is the data distribution gap between training and inference stage, 
% where the training data is clean sentence-level pairs while the test data is noisy dialogue-level conversations.
where the training data are clean sentence-level pairs while the test data are noisy dialogue-level conversations.

To bridge the distribution gap, the first step is to generate perturbation examples based on training instances.
% The data generation mainly consists of two steps, with the first step to obtain sub-documents with cross-sentence context and the second step to generate examples with word perturbations within sub-documents.
The data generation mainly consists of two steps. The first step is to obtain sub-documents with cross-sentence context, and the second step is to generate examples with word perturbations within sub-documents.
% 图二的左边部分展示了一个完整的流程
Figure \ref{pdf_model_a} shows a complete process.

\noindent\textbf{Cross-sentence Context} It is difficult to acquire dialog-level parallel training data.
As an alternative approach, we use parallel document data to catch dependencies across sentences.

Formally, let $x_d=\{x^{(1)}, x^{(2)},\cdots, x^{(M)}\}$ be a source-language document containing $M$ source sentences. % We use $L_x^{(i)}$ to denote the sentence length of $X^{(i)}$. 
And $y_d=\{y^{(1)}, y^{(2)},\cdots, y^{(M)}\}$ is the corresponding target-language document containing the same number of sentences as that of the source document.
To get more context information, we randomly sample consecutive sub-document pairs $(x_d,y_d)$ of $N$ sentences (i.e., snippet pairs from aligned documents). We set $N \in [1, 10]$ in this paper.
% , with the probability of 2.5\% to improve training efficiency.

We use a special token $<$$\texttt{sep}$$>$ as the separator to concatenate sentences into a parallel sub-document $\{(x_d,y_d)\}$, as shown in Figure \ref{pdf_model_a}. 
% To create perturbation training data for dialogue translation, we also introduce the huge amount of parallel sentences ($N=1$) into the parallel sub-document training data constructed according to the way mentioned above. 

\noindent\textbf{Contextual Perturbation}
We then consider generating perturbation example $x_d'$ from $x_d$ with respect to sub-document context.
For \droppro, \droppun and \typo, we build a Chinese pronoun table $\text{T}_{\droppro}$, a common punctuation table $\text{T}_{\droppun}$ and a Chinese homophone table $\text{T}_{\typo}$ respectively.

For \droppro and \droppun, we traverse source sentences of $x_d$, discard pronouns/punctuation in these sentences with a probability of 30\% and record deletion positions with corresponding labels (see details below); to construct a typo, we choose a word with a probability of 1\%, of which 80\% is replaced with one of its homophones according to $\text{T}_{\typo}$ and 20\% is replaced with another random word. 
We determine these percentages by observing the generated perturbation data.
For annotation labels, we tag correct words with 0, words of \typo with 1, \droppro words with 2 and \droppun words with 3.  

Finally we get $x_d$, $x_d'$ and their corresponding label sequences $\ell_x$, $\ell_x'$. $\ell_x$ is a sequence of all 0s.

\subsection{\method Base Models}
With the created training data, we first introduce two methods for \method as our strong baselines, which will be elaborated here for model comparison. 

\noindent\textbf{\mrepair}
A natural way for \method is to train a dialog repair model to transform dialogue inputs into forms that an ordinary NMT system can deal with. \mrepair involves training a repair model to transform $x_d'$ to $x_d$ and a clean translation model that translates $x_d$ to $y_d$.
As a pipeline method, \mrepair may suffer from error propagation.

\noindent\textbf{\mrobust}
We extend the robust NMT \cite{DBLP:conf/acl/LiuTMCZ18} to dialogue-level translation.  Specifically, we take both the original $(x_d,y_d)$ and the perturbated $(x_d', y_d)$ bilingual pairs as training instances. 
So the model is more resilient on dialogue translation. During the inference stage, the robust model directly translates raw inputs into the target language. 

\subsection{\mmlt}
\mrobust has the potential to handle translation problems caused by noisy dialogue inputs. However, the internal mechanism is rather implicit and in a black box. Therefore, the improvement is limited, and it is not easy to analyze the improvement. 
To address this issue, we introduce a context-aware multi-task learning method \mmlt for \method.  

As shown in ③ of Figure \ref{pdf_model_b}, the only difference is that we have a contextual labeling module based on the encoder. We denote the final layer output of the Transformer encoder as $H$. For each token $h_i$ in $H=(h_1, h_2, ..., h_m)$, the probability of contextual labeling is defined as:
\begin{equation}
P(p_i=j|X)=softmax(W\cdot h_i+b)[j]
\end{equation}
where $X=(x_1, x_2, ..., x_m)$ is the input sequence, $P(p_i=j|X)$ is the conditional probability that token $x_i$ is labeled as $j$ ($j \in {0,1,2,3}$ as defined above).

Here we make the labeling module as simple as possible, so that the Transformer encoder can behave like BERT \cite{devlin2019bert}, learning more information related to perturbation and guiding the decoder to find desirable translations.

During the training phrase, the model takes $(x_d, x_d', \ell_x, \ell_x', y_d)$ as the training data.
The learning process is driven by optimizing two objectives, corresponding to sequence labeling as auxiliary loss ($\mathcal{L}_{SL}$) and machine translation as the primary loss ($\mathcal{L}_{MT}$) in a multi-task learning framework.
\vspace*{-0.5\baselineskip}
\begin{align}
\mathcal{L}_{SL} = -log(P(\ell_x|x_d)+P(\ell_x'|x_d')) \\
%\end{equation}
%\begin{equation}
\mathcal{L}_{MT} = -log(P(y_d|x_d)+P(y_d|x_d'))
\end{align}
The two objective are linearly combined as the overall objective in learning.
\begin{equation}
\mathcal{L} = \mathcal{L}_{MT} + \lambda\cdot\mathcal{L}_{SL}
\end{equation}
$\lambda$ is coefficient. During experiments, we set as follows according the best practice:
\vspace*{-0.5\baselineskip}
\begin{equation}
\lambda=max(1.0-\frac{update\_num}{10^5}, 0.2)
\end{equation}
where $update\_num$ is the number of updating steps during training.
% At the inference stage, \mmlt takes the raw dialogue $\hat{d}_x'$ as the input and generates the target translation $\hat{\ell}_x'$ and $\hat{y}_d$.

% 能否加一段为什么用multi-task learning的理由，为什么要做context labeling，这个对dialogue translation有什么帮助？
% 我们引入多任务学习的原因有两点：1.labeling的表现一定程度反应了模型对于包含上文提到现象的句子的理解，相比于repair模型和robust模型有一定的可解释行。2.对于labeling的训练可以视为一个基于BERT的序列标注任务，显式的引导可以让encoder学习到更多我们标注的信息。
% what's more， 我们的method不仅限于对话翻译或者是中英翻译。我们可以根据具体的问题，设计自动方法生成标记信息，用来指导模型学习。
We introduce multi-task learning for two reasons: 1) The labeling performance reflects the model's understanding of sentences containing the mentioned phenomena. 
% It has a certain level of interpretability as compared to \mrepair and \mrobust.
2) Contextual Labeling can be seen as a pre-training process based on the BERT-like model, and explicit guidance can enable the encoder to learn more about the information we annotate.
% Moreover, our method is not limited to Chinese-English or dialogue translation. We can guide models by designing automatic methods to generate annotation according to specific problems.

\subsection{Modeling  Dialogue Context}
The modes for exploring dialogue context during decoding can be divided into offline and online.
For the offline setting, all sentences in a dialogue are concatenated one by one with $<$\texttt{sep}$>$. The concatenated sequence is translated, and the target translation for each sentence can be easily detected according to the separator $<$\texttt{sep}$>$.

The offline mode can be used for dialogue translation where the entire source  dialogue has already been available before translation (e.g., movie subtitles). However, we continuously get new source sentences for online chat and need to generate corresponding translations immediately. We refer to this mode as the online setting. 

We experiment with two online methods. One is \textit{online-cut} where the current sentence is concatenated to the previous context with the separator $<$\texttt{sep}$>$. The trained \method model then translates the concatenated sequence and the last target segment is used as the translation for the current source sentence. 
The other is \textit{online-fd}. \textit{Online-fd} is a force decoding method. It forces the decoder to use translated history and continues decoding instead of re-translating the entire concatenated sequence. \textit{Online-fd} brings more consistent translation.

% However, this method has two main disadvantages: 1) we have only previous context, the following context is lost; 2. we only use part of the whole translation, and it hurts the coherence of the whole translation.

%\section{Benchmarking Dialogue Translation}
%\label{sec:testsets}
%\input{04-testsets.tex}

\section{Experiments}
\label{sec:exp}
\subsection{Test Set}
For better evaluation of \method, we create a Chinese-English test set covering all issues discussed above based on the corpus we analyze in the second section. 
\begin{table}
\centering
\small
%\scriptsize
\begin{tabular}{l|c}
\hline
\textbf{Item}   & \textbf{Count} \\ \hline
\#dialogues & 300             \\
\#sentence pairs  & 1,931           \\
\#total tokens    & 19,155/15,976   \\
\#average tokens  & 9.92/8.27       \\
\#\droppro      & 299             \\
\#\droppun      & 542             \\
\#\typo    & 203             \\ \hline
\end{tabular}
\caption
{Statistics on the test set. ``/'' denote numbers in Chinese and English separately.}
\label{table_statistics}
\end{table}
Statistics on the built test set are displayed in Table \ref{table_statistics}. 
%We show the number of dialogues, bilingual pairs, \droppro/\droppro/\typo, total tokens, and the average number of tokens per sentence. 
Building such a test set is hard and time-consuming as we need to perform manual selection, translation and annotation. 

As for translation quality evaluation, we use other metrics in addition to BLEU.
% We describe here how we evaluate dialogue translation quality on this test set in addition to using n-gram based metrics such as BLEU.
% As shown in Table \ref{table_general}, we will test from the overall and details aspects. 
For \droppun and \typo, we evaluate BLEU scores on sentences containing missing punctuation or typos according to the annotation information. 
%As for \droppro, because we have annotated the missing pronoun in each \droppro sentence and its corresponding English translation during construction, and almost each sentence in our test set containing \droppro has no other explicit pronouns, we can judge if a tested model is able to successfully recover and translate dropped pronouns by checking whether dropped pronouns exist in translation. We evaluate the translation quality of the model for \droppro phenomenon by the percentage of correctly recovering and translating the dropped pronouns.
As for \droppro, we evaluate the translation quality by the percentage of correctly recovering and translating the dropped pronouns.

% All the results in Table \ref{table_general} are under the offline setting with all the context in the dialogue can be used.
%All the results in Table \ref{table_general} are under the offline setting while all the context in the dialogue can be used.
%We also evaluate results in online setting with different lengths of context in Table \ref{table_compare_context} and Figure \ref{pdf_offline}.

\subsection{Settings}
We adopt the Chinese-English corpus from WMT2020\footnote{This corpus includes News Commentary, Wiki Titles, UN Parallel Corpus, CCMT Corpus, WikiMatrix and Back-translated news.}, with about 48M sentence pairs, as our bilingual training data $D$. We select newstest2019 as the development set. After splicing, we get $D_{doc}$ with 1.2M pairs and corresponding perturbated dataset $D'$ and $D_{doc}'$ with 48M and 1.2M pairs respectively.

We use byte pair encoding compression algorithm (BPE) \cite{sennrich2016neural} to process all these data and limit the number of merge operations to a maximum of 30K.
In our studies, all translation models are Transformer-big, including 6 layers for both encoders and decoders, 1024 dimensions for model, 4096 dimensions for FFN layers and 16 heads for attention. 

During training, we use label smoothing = 0.1 \cite{szegedy2016rethinking}, attention dropout = 0.1 and dropout \cite{Hinton2012ImprovingNN} with a rate of 0.3 for all other layers. We use Adam \cite{adam} to train the NMT models. ${\beta} 1$ and ${\beta} 2$ of Adam are set to 0.9 and 0.98, the learning rate is set to 0.0005, and gradient norm 5.
The models are trained with a batch size of 32,000 tokens on 8 Tesla V100 GPUs during training. During decoding, we employ beam search algorithm and set the beam size to 5.
We use sacrebleu \cite{post-2018-call} to calculate uncased BLEU-4 \cite{Papineni2001BleuAM}.
% It is worth noting that some models will supplement the punctuation at the end of a translation while others not. 
% It is worth noting that some models will automatically add the missing punctuation at the end of the sentence.
% It will cause a significant BLEU score difference, but have no impact on the understanding of the translation. 
%So for all the results, we uniformly remove punctuation at the end of translations.
% So we remove the punctuation at the end in each sentence.

% \subsection{Baselines}
% Because there is basically no work related to dialogue translation before, we compare our model with baselines including base, repair and robust model.

% \noindent\textbf{base}
% The base model is the original Transformer-big trained with $D$ and $D_{concat}$. With the help of $D_{concat}$, the base model can translate single sentence and spliced sentences at the same time.

% \noindent\textbf{repair}
% The repair method is a pipeline of a repair model and a base model mentioned above. The repair model recover annotated input to raw and is trained with the source-side part of ($D^{annot}$, $D_{concat}^{annot}$) and ($D$, $D_{concat}$). The base model then reads the repaired input and translate normally.

% \noindent\textbf{robust}
% Similarly, the robust model is trained with $D$, $D_{concat}$, $D^{annot}$ and $D_{concat}^{annot}$. We use both annotated and raw corpus at the same time. We hope that the robust model reads annotated input and produces correct output.

\begin{table}[]
\centering
\small
\resizebox{\linewidth}{!}{
\begin{tabular}{l|c|ccc}
\hline
\multirow{2}{*}{Methods} & Overall       & \multicolumn{3}{c}{Details}                      \\
                         & BLEU          & \droppro          & \droppun      & \typo    \\ \hline
\mbase                     & 32.7          & 26.09\%          & 28.2          & 24.0 \\
\mrepair                   & 34.0          & 29.77\%          & 31.2          & 27.4          \\
\mrobust                   & 34.1          & 45.48\%          & 33.0          & \textbf{28.8}          \\
\mmlt              & \textbf{35.9} & \textbf{47.16\%} & \textbf{34.3} & 28.7          \\ \hline\hline
GOLD+\mbase                & \textbf{36.8} & \textbf{97.32\%} & \textbf{34.6} & \textbf{36.8}   \\ \hline
\end{tabular}}
\caption{Experiment results on our constructed dialogue translation test set in offline setting. The GOLD+\mbase represents translations of completely correct inputs (without \droppro, \droppun or \typo) using \mbase model, which is used to show the oracle results with Transformer on the test set.}
\label{table_general}
\end{table}

\begin{table}[]
\resizebox{\linewidth}{!}{
\begin{tabular}{l|l|lll}
\hline
\multirow{2}{*}{Methods} & Overall    & \multicolumn{3}{c}{Details}                \\
                         & BLEU       & \droppro              & \droppun        & \typo      \\ \hline
\mbase                     & 32.8(+0.1) & 19.06\%(-7.03\%) & 28.1(-0.1) & 22.3(-1.7) \\
\mrepair                   & 33.8(-0.2) & 24.75\%(-5.02\%) & 32.0(+0.8) & 28.3(+0.9) \\
\mrobust      & 34.2(+0.1)          & \textbf{36.79\%(-8.69\%)} & 32.7(-0.3)          & \textbf{28.9(-0.5)} \\
\mmlt   & \textbf{35.3(-0.6)} & 34.78\%(-12.38\%)         & \textbf{34.3(-0.0)} & 28.6(-0.1)          \\ \hline\hline
GOLD+\mbase   & \textbf{37.1(+0.3)} & \textbf{96.66\%(-0.66\%)} & \textbf{35.3(+0.7)} & \textbf{35.9(-0.9)} \\ \hline
\end{tabular}}
\caption{Results on our constructed dialogue translation test set in online setting at the sentence level.}
\label{table_compare_context}
\end{table}

\begin{figure}[t]
\centering
\includegraphics[scale=0.5]{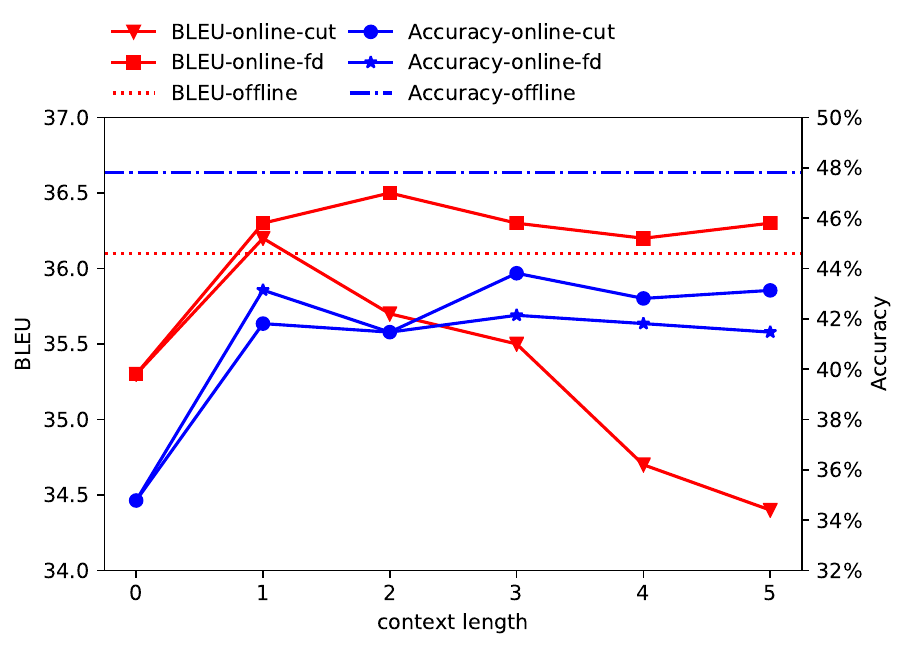}
\caption
{Overall BLEU and \droppro recovery performance (Accuracy) of \mmlt with different context length. Dash lines are the offline results.}
\label{pdf_offline}
\end{figure}

\subsection{Results of Offline Setting}
The offline mode aims at using the entire source dialogue for translation.
We experiment with all the methods in the offline setting, and the results are shown in Table \ref{table_general}. \mbase is a Transformer-big model trained with $D$ and $D_{doc}$. 
GOLD+\mbase represents the oracle result on this test set. 
We can see that \mmlt has achieved the best results, reducing the gap between $test_{wrong}$ and $test_{gold}$ from 4.1 to 0.9. Compared with \mrobust and \mmlt, \mrepair performs relatively poorly. We believe that this is due to the error propagation caused by the pipeline.

From the specific indicators, 
%we can get the following conclusions:
we can draw the following conclusions:
% \begin{itemize}
%     \item \typo has a very obvious impact on BLEU, and the gap between \mbase and GOLD+\mbase is more than 12 points. Although all methods have a certain effect on \typo, they produce very different results from the best one. We will further analyze the problem in the Analysis section.
%     \item The recovery of \droppro is a relatively difficult task. Although compared with \mbase, the current best result of 47.16\% has been greatly improved, but is still far away from 97.32\%.
%     \item \droppun seems to be a relatively easy task for each method to address. Basically all methods can bring good improvements. For \mmlt, the BLEU score can even reach the level of $test_{gold}$.
% \end{itemize}
1) \typo has a very obvious impact on BLEU, and the gap between \mbase and GOLD+\mbase is more than 12 points; 2) The recovery of \droppro is a relatively difficult task. Although compared with \mbase, the current best result of 47.16\% has been greatly improved, but is still far away from the golden result 97.32\%; 3) \droppun seems to be a relatively easy task for each method to address.

\subsection{Results of
 Online Setting }
The online mode only makes use of previous context during translation.
An extreme situation of online setting is that there is no context, that is, sentence-level translation.
We show the results of all the methods on the test set at the sentence level in Table \ref{table_compare_context}.
Despite the lack of context, our approaches can still bring general benefits.
We find that \droppro relies heavily on context, especially for \mmlt, where the absence of context results in a 12.38\% drop in performance. 
% This is in line with our expectations, as in many cases dropped pronouns appear in context.
This is in line with our expectations, as in many cases  machine translation system heavily depends on context to fulfill the dropped pronouns.
% In contrast, context has little effect on \droppun, as the semantics of the current sentence is generally sufficient to split itself. For \typo, context has a slight positive effect.

% We further experiment how context length affect \method.
We further experiment on how context lengths can affect \method.
The results are shown in Figure \ref{pdf_offline}. In the \textit{online-cut} setting, we can see that using previous few sentences as context may improve overall BLEU score, but continuously adding more preceding texts will lead to a continuous decline. \textit{Online-fd} performs well because using historical translation records to continue decoding can bring more consistent translation results.
For the recovery accuracy of \droppro, \textit{online-cut} is better than \textit{online-fd} in contrast, because forced decoding may cause wrong pronoun transmission.
%The preceding sentences can bring a significant improvement, but when it reaches a certain length, it does not improve anymore. 

%Both online modes do not offer as accurate \droppro recovery as those of offline mode because they have only previous context, and the following context is lost.

\section{Analysis}
\label{sec:analysis}
\begin{table}[]
\centering
\small
\begin{tabular}{|c|c|c|c|c|}
\hline
Data                   &       & Precison      & Recall        & F1            \\ \hline
\multirow{3}{*}{validation} & \droppro   & \textbf{61.3} & \textbf{48.7} & \textbf{54.3} \\ \cline{2-5} 
                       & \droppun   & 80.0          & 63.6          & 70.9          \\ \cline{2-5} 
                       & \typo & \textbf{85.3} & \textbf{64.2} & \textbf{73.2} \\ \hline\hline
\multirow{3}{*}{test}  & \droppro   & 48.6          & 32.2          & 38.8          \\ \cline{2-5} 
                       & \droppun   & \textbf{96.6} & \textbf{87.9} & \textbf{92.1} \\ \cline{2-5} 
                       & \typo & 83.3          & 31.0           & 45.2           \\ \hline
\end{tabular}
\caption{Labeling performance on the validation/test set.}
\label{table_error-detection}
\end{table}

\subsection{Labeling Performance}
To better understand how our proposed \mmlt make sense, we calculate the labeling performance on both validation and test set. Table \ref{table_error-detection} shows the overall performance. The validation set follows the same processing progress of training data, while the test set is the real dialogue data set built manually. 

The proposed model obtains 54.3\% F1 score on the validation set for \droppro, 70.9\% for \droppun, and 73.2\% for \typo. When testing on the real test data, the performance on \droppro has declined a lot because of the difference between synthetic training/validation data and real test data. 
%The performance on the test set for \droppun is even better. We think this is because sentences of the test set are short and relatively simple to split.
% 特别需要注意的是，在测试集上typos的成绩特别低，f1只有. 这是由于极地的召回率，只有3.8，接近于0。这表明了我们的模型虽然在开发集上能很好得检测错别字，但在测试集上基本完全检测不到。这可能是因为我们通过自动方法生成的错别字和真实分布的错别字存在过大的差别。我们分析了测试集中的错别字和自动生成的，发现。。。
Especially noteworthy is the fact that F1 score of \typo drops the most, reaching 26\%, because of its low recall. It may be due to the considerable difference between the typos generated by our automatic method and the actual distribution. 

\begin{table}[]
\resizebox{\linewidth}{!}{
\begin{tabular}{|l|l|l|}
\hline
    & zh         & en                                         \\ \hline
(1) &
  \begin{tabular}[c]{@{}l@{}}艾丽最近怎么样了\\ 她已经不在我的律所了\\ 什么 \textbf{(她/she)}为什么走了\\ \textbf{(她/she)}开了自己的律所\end{tabular} &
  {\begin{tabular}[c]{@{}l@{}}What's going on with Ellie?\\ She is no longer in my law firm.\\ What, why \color[HTML]{CB0000}\textbf{are you}\color[HTML]{000000}/\color[HTML]{3166FF}\textbf{is she }\color[HTML]{000000} going?\\ \color[HTML]{CB0000}\textbf{I open my}\color[HTML]{000000}/\color[HTML]{3166FF}\textbf{She opens her }\color[HTML]{000000} own law firm.\end{tabular}} \\ \hline
(2) & 琼斯 \textbf{(我/I)}问你件事 & Jones \color[HTML]{CB0000}\textbf{asked}\color[HTML]{000000}/\color[HTML]{3166FF}\textbf{, I want to ask }\color[HTML]{000000} you something. \\ \hline
(3) &
  \begin{tabular}[c]{@{}l@{}}他上次帮我私下搞\\ \textbf{(他/He)}差点工作都丢了\end{tabular} &
  {\begin{tabular}[c]{@{}l@{}}He helped me out in private last time.\\ \color[HTML]{CB0000}\textbf{I}\color[HTML]{000000}/\color[HTML]{3166FF}\textbf{He }\color[HTML]{000000} nearly lost \color[HTML]{CB0000}\textbf{my}\color[HTML]{000000}/\color[HTML]{3166FF}\textbf{his }\color[HTML]{000000} job.\end{tabular}} \\ \hline
\end{tabular}}
\caption{Examples of \droppro recovery errors.}
\label{table_dpr_recovery_error}
\end{table}

\begin{figure}[t]
\centering
\includegraphics[scale=0.35]{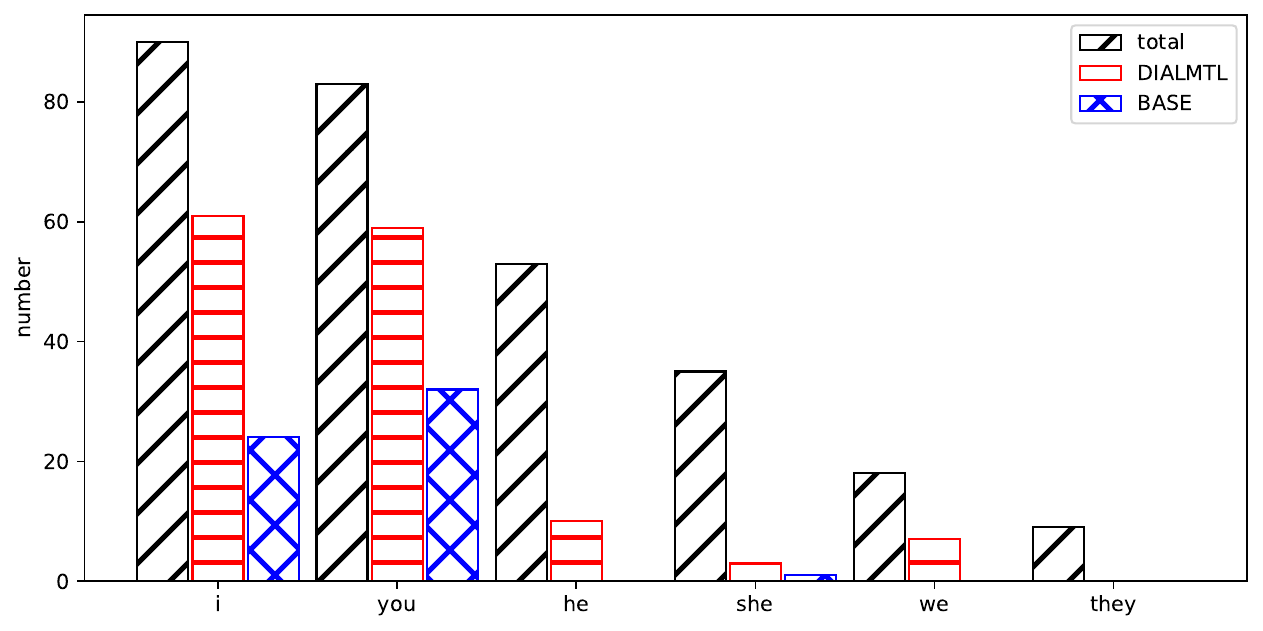}
\caption
{\droppro recovery performance of \mbase and contextual \mmlt. Total means the total number of occurrence of corresponding pronouns in the test set. We ignore pronouns with a total occurrence number less than 5.}
\label{pdf_recoverydpr}
\end{figure}

\subsection{Effects of Pronoun Correcting}
We further explore the auto-correction of specific pronouns. As shown in Figure \ref{pdf_recoverydpr}, we can find that pronouns such as I/you, which occur mostly in the corpus, generally have a higher recovery success rate. We believe this is due to the data imbalance. Compared with \mbase, \mmlt has a much better performance.
% In particular, we notice that the recovery success rate of \mbase on pronouns except i/you is almost zero.
While \droppro recovery accuracy has been improved, it still has not achieved 50\%. 
The most common error is that the model does not capture any context or captures previous inappropriate context. We summarize frequently-occurring recovery errors in Table \ref{table_dpr_recovery_error}.
% The most common error is model not capturing, or capturing the wrong previous context.
 
% As shown in (1), when asking ``为什么走了/why going'', we refer to ``艾丽/Ellie'' mentioned above. However, the model automatically fills the omission with ``you", which causes a chain reaction below, using ``I" in the next sentence. 
% This problem is the most common one and needs to be solved. We think the key point is the lack of real dialogue data, making the model difficult to capture relationships between real interlocutors only by continuous sentences. 
%(2) is an error caused by \droppun. ``琼斯/Jones" is mistaken as the pronoun here. 
% There are also some difficult scenarios, as shown in (3). Only looking at these two sentences, ``我/I" or ``他/He" can both make sense. So more context is needed, including context outside the dialogue.

\section{Related Work}
\label{sec:relatedwork}
% Our work mainly aims at dialogue translation. Up to now, there exists few works in this area except extracting bilingual dialogue from some existing subtitle data sets. The \droppro problem mentioned in this paper is related to dropped pronouns recovery. There have existed some studies on how to recover pronouns and improve translation quality at the same time. In order to solve the problems in dialogue translation, we choose to automatically generate noisy data from a large number of parallel corpus, and add error detection task for multi-task training. This is related to  robust training.
Our work is related with both dialogue translation % dropped pronoun recovery 
and robust training.
% 对话翻译的相关工作
% Contextual Neural Model for Translating Bilingual Multi-Speaker Conversations

\noindent\textbf{Dialogue Translation}

There has been some work on building bilingual dialogue data sets for the translation task in recent years. \citet{wang2016automatic} propose a novel approach to automatically construct parallel discourse corpus for dialogue machine translation and release around 100K parallel discourse data with manual speaker and dialogue boundary annotation.
\citet{maruf2018contextual} propose the task of translating Bilingual Multi-Speaker Conversations. They introduce datasets extracted from Europarl and Opensubtitles and explore how to exploit both source and target-side conversation histories.
\citet{bawden2019diabla} present a new English-French test set for evaluating of Machine Translation (MT) for informal, written bilingual dialogue.
Recently WMT2020 has also proposed a new shared task - machine translation for chats,\footnote{http://www.statmt.org/wmt20/chat-task.html} focusing on bilingual customer support chats \cite{farajian2020findings}.

% DPr的相关工作
% Recovering dropped pronouns from Chinese text messages
% SMT的一堆工作 NMT的一些工作
% \noindent\textbf{Dropped Pronoun Recovery}
% \citet{yang2015recovering} are the first to work on recovering dropped pronouns in Chinese text messages. \citet{giannella2017dropped} propose to employ a linear-chain CRF classifier to deal with this problem. \citet{zhang2019neural} introduce a multi-layer perceptron model for the first time. \citet{DBLP:conf/naacl/YangT0GGX19} present a novel end-to-end model to recover dropped pronouns by modeling the referents of dropped pronouns.
% There has also been some works\cite{wang2016novel, wang2016dropped, wang2018translating, wang2018learning} that try to recover dropped pronouns to help NMT when translating from pro-drop languages to non-pro-drop languages. However, although dialogue datasets are used, their work often does not consider problems in the context of dialogue.

\noindent\textbf{Robust Training}

Neural models have been usually affected by noisy issues. Many efforts \cite{li2017robust, sperber2017toward, vaibhav2019improving,yang2020towards} focus on data augmentation to alleviate the problem by adding synthetic noise to the training set. However, generating noise has always been a challenge, as natural noise is always more diversified than artificially constructed noise \cite{belinkov2018synthetic, anastasopoulos2019analysis, anastasopoulos2019neural}.

% 上下文相关工作
% 鲁棒性相关工作

\section{Conclusions}
\label{sec:conclusions}
In this paper, we manually analyze challenges in dialogue translation and detect three main problems. In order to tackle these issues, we propose a multi-task learning method with contextual labeling. For deep evaluation, we construct dialogues with translation and detailed annotations as a benchmark test set. Our proposed model achieves substantial improvements over the baselines. What is more, we further analyze the performance of contextual labeling and pronoun recovery errors. 
% We leave sentence segment errors and how to better use context in online setting for future work.

\section*{Acknowledgments}
We thank the bilingual speakers for test set construction, and the anonymous reviewers for suggestions. Deyi Xiong is partially supported by the Natural Science Foundation of Tianjin (Grant No. 19JCZDJC31400) and the Royal Society (London) (NAF\verb|\|R1\verb|\|180122).

\bibliography{anthology,custom}
\bibliographystyle{acl_natbib}

%\appendix

%\section{Example Appendix}
%\label{sec:appendix}

%This is an appendix.

\end{CJK}
\end{document}